\RequirePackage{fix-cm}
\documentclass[smallextended]{svjour3}       
\smartqed  

\def\c{\ensuremath{\mathbf{c}}}
\def\cau{\ensuremath{\mathbf{c}}}
\def\e{\ensuremath{\mathbf{e}}}
\def\g{\ensuremath{\mathbf{g}}}
\def\f{\ensuremath{\mathbf{f}}}

\usepackage{tikz}
\usetikzlibrary{arrows,automata,positioning}
\usetikzlibrary{mindmap,trees}
\definecolor{shadecolor}{rgb}{0.2,0.9,0.2}
\definecolor{green}{rgb}{0,0.3,0}
\definecolor{OliveGreen}{rgb}{0,0.2,0}
\usepackage{caption}
\usepackage{graphicx}
\usepackage{amsmath}
\usepackage{bm}
\usepackage{amscd,latexsym}
\usepackage{amsfonts}   
\usepackage{amssymb,amsthm}
\usepackage{mathrsfs}
\usepackage[section]{placeins}
\usepackage[misc]{ifsym}
\usepackage{array}
\usepackage{cite}
\usepackage[colorlinks=true,citecolor=blue,urlcolor=blue]{hyperref}
\usepackage{multirow}
\usepackage{makecell}
\usepackage{booktabs}
\DeclareMathSymbol{\smin}{\mathbin}{AMSa}{"39}
\begin{document}

\title{Inference over Radiative Transfer Models using Variational and Expectation Maximization Methods
\thanks{Research partly funded by the ERC under the ERC-CoG-2014 project (grant agreement 647423) and the ERC-SyG-2019 USMILE project (grant agreement 855187), the Ministerio de Ciencia e Innovación project PID2019-106827GB-I00/AEI/10.13039/501100011033, and the projects TEC2016-77741-R, DPI2017-89867-C2-2-R, RTI2018-096765-A-I00, PID2019-109026RB-I00 and PID2019-105032GB-I00.}}

\author{Daniel Heestermans Svendsen \and
        Daniel Hern\'andez-Lobato   \and
        Luca Martino \and
        Valero Laparra \and
        \'Alvaro Moreno-Mart\'inez \and
        Gustau Camps-Valls
}

\institute{\Letter ~ Daniel Heestermans Svendsen \at
              \email{daniel.svendsen@uv.es} \\
              Image Processing Laboratory (IPL), Universitat de Val\`encia, Spain
           \and
           Daniel Hern\'andez-Lobato \at
              Dep. Computer Science, Universidad Aut\'onoma de Madrid, Spain
           \and
           Luca Martino \at
              Dep. Signal Theory and Telecommunications, Universidad Rey Juan Carlos de Madrid
           \and
           Valero Laparra \and
           Alvaro Moreno \and
           Gustau Camps-Valls \at
              Image Processing Laboratory (IPL), Universitat de Val\`encia, Spain
}

\date{2021 Springer. Published in Machine Learning (2021). DOI: 10.1007/s10994-021-05999-4}

\maketitle

\begin{abstract}
    Earth observation from satellites offers the possibility to monitor our planet with unprecedented accuracy. 
    Radiative transfer models (RTMs) encode the energy transfer through the atmosphere, 
    and are used to model and understand the Earth system, as well as to estimate the parameters that describe the status of the Earth from satellite observations by inverse modeling. 
    However, performing inference over such simulators is a challenging problem. RTMs are nonlinear, non-differentiable and computationally costly codes, which adds a high level of difficulty in inference. 
    In this paper, we introduce two computational techniques to infer not only point estimates of biophysical parameters but also their joint distribution. 
    One of them is based on a variational autoencoder approach and the second one is based on a Monte Carlo Expectation Maximization (MCEM) scheme. We compare and discuss benefits and drawbacks of each approach. We also provide numerical comparisons in synthetic simulations and the real PROSAIL model, a popular RTM that combines land vegetation leaf and canopy modeling. We analyze the performance of the two approaches for modeling and inferring the distribution of three key biophysical parameters for quantifying the terrestrial biosphere. 

\end{abstract}

\section{Introduction} \label{see:intro}

In many areas of science and engineering, systems are analyzed by running computer code simulations, which act as convenient approximations to reality. Depending on the body of literature, they are known as physics-based, processed-oriented and mechanistic models, or simply just {\em simulators} \cite{Santer03,Wescott13}. Simulators  
are ubiquitous in physics, brain, social, Earth and climate sciences \cite{Raissi17,sandberg2013feasibility,amogape20}.
Model simulations are needed to understand system behaviour, but also to perform counterfactual studies. 

In Earth sciences the use of simulators is of paramount importance. Earth observation (EO) from  airborne and satellite remote sensing platforms along with  in-situ observations play a fundamental role in monitoring our planet \cite{lillesand08,Liang08,rodgers00}. 
Remote sensing simulators of the involved processes are known as {\em radiative transfer models} (RTMs). 
These models describe the complex interactions of scattering and absorption of radiation with the constituents of the atmosphere, water, vegetation and soils. RTMs are useful because they allow us to translate (map) a set of parameter\footnote{ {Note that both the field of remote sensing parameter retrieval and the field of Bayesian inference make heavy use of the word \textit{parameter}. In the former case it refers to the physical parameters over which we are inferring distributions in this work, and in the latter case it refers to distribution parameters (mean, covariance, etc.). The meaning is always clear from context, however. } } values describing the state of soil, leaf, canopy and atmosphere to at-sensor reflectance or radiance. Such simulations allow for modeling, understanding, and predicting parameters related to the state of the land cover, water bodies and atmosphere.

While modeling and characterizing the involved processes is key, in practice one is typically interested in solving the so-called {\em inverse problem}
; that is, for example, inferring the set of atmospheric or canopy biophysical properties, so that the computed reﬂectances best ﬁt the remotely sensed ones \cite{tarantola2005inverse,jointgp,zurita2015visualizing}. The problem of inverting the forward model is in general highly {\em ill-posed} \cite{combal2003retrieval}: 
Different sets of parameter values can map into the same reflectance, thus making it difficult to recover the true set of parameters given a remotely sensed reflectance.
This issue has been largely reported in the literature \cite{Verstraete96,Knyazikhin99,Liang08} and is, together with the  complexity and computational cost of the RTMs as well as the scarcity of labeled data, the main reasons why the inverse problem is a difficult and unresolved one.

Many methods have been proposed for model inversion. Early approaches considered minimizing the (e.g. least squares) error between observations and model simulations stored in big Look-Up-Tables (LUTs). {Comparing each observed spectrum with all spectra stored in the LUT proved impractical, leading to gradient-descent techniques in combination with emulators to be proposed in the literature} \cite{Jorge19linearagape}.   
More advanced approaches have recently exploited machine learning regression algorithms, such as random forests \cite{liang2015estimation,campos2018global}, neural networks \cite{baret2007lai,djamai2019validation} and Gaussian processes \cite{amogape20,jointgp,PhysicsAwaregp,CampsValls19nsr,campsvalls16grsm} to achieve improved multidimensional interpolation capabilities. 
Treating the inverse problem purely as a regression problem, however, only leads to point-wise estimates, and not a joint probability distribution of the parameters. {We argue in this work that when two or more physical variable configurations result in the same spectrum, a conventional inversion method will perform poorly. Such methods that try to predict the cause from the effect will simply predict something in between the possible causes that gave rise to the effect. This problem corresponds to having a multimodal posterior and can be addressed with one of the proposed probabilistic frameworks in this paper.}

A Bayesian formalism is useful in order to (1) generate probability density functions (PDFs) of the parameters, and hence account for all moments and uncertainty on the retrieved parameters \cite{pinty2011exploiting,lewis2012earth},  (2) to incorporate constraints in the form of a-priori parameter distributions, which are often subject of intense debate in the literature \cite{combal2003retrieval,baret2007lai,atzberger2012spatially}, and (3) to overcome the limited potential of iterative steepest-descent optimization procedures to locate globally optimal solutions \cite{bacour2002reliability,zhang2005estimating}. {Furthermore, deriving a generative model provides a straightforward way to perfrom outlier detection, by measuring the probability of the observed data under the fitted model.}

Given some vector of physical causes $\cau$ (atmospheric or canopy properties), the forward RTM model induces a likelihood function  $p(\e|\cau)$, which links the causes with the physical effects $\e$ (reflectance spectra). In this work, we address a general problem: Learning the distribution of the physical parameters or \textit{causes}, instead of only providing a pointwise estimation of these parameters (by statistical or numerical inversion).  Provided a dataset of observed effects $\e'$, our goal is twofold: learning the marginal density $p(\cau)$ and obtaining an approximation of the conditional distribution $p(\cau|\e')$, which in a Bayesian setting represents the posterior density of the causes given the effects. Note that $p(\cau|\e')$ also represents a probabilistic inverse model, i.e., given $\e'$ we can obtain a prediction of the causes $\cau$ and related uncertainty measures. {Probabilistic inverse modelling, although not so widely used in Remote Sensing applications, has proven to be a powerful tool, providing more general (and hence potentially more valuable) solutions than point-wise approaches, and can help in better understanding the problem itself \cite{zhang2005estimating,coccia2015creating,ma2017uncertainty}. }

Since RTMs are generally complex, non-differentiable (i.e. having non-analytical Jacobian) and computationally costly models, mathematical tractability is typically compromised, especially when the aim is to combine RTMs and Bayesian methods. 
Here, we propose and compare two different approaches which allows us to infer parameters for a non-differentiable simulator. One approach is based on Monte Carlo Expectation Maximization (MCEM) \cite{wei1990monte} and the other is based on Variational Autoencoders (VAEs) \cite{kingma2013auto}.  
We will show that each approach has different pros and cons. While the MCEM approach is mathematically elegant, flexible and has good convergence properties, its application in practice is computationally demanding. On the other hand the proposal based on a simple version of VAE obtains good results and is fast, yet it is not able to describe multimodal distributions. While possible, its extension for multimodal distributions makes the approach more complicated, reducing the good computational properties (see, \emph{e.g.}, \cite{mescheder2017adversarial}). We illustrate these properties in several toy examples of varying sample sizes and complexity, as well as with the PROSAIL RTM \cite{baret1992modeled}.  {PROSAIL is the combination of the PROSPECT \cite{jacquemoud1990prospect} leaf optical properties model and the SAIL \cite{verhoef1984light} (Scattering by Arbitrary Inclined Leaves) canopy reflectance model}. In particular, we compare the approaches for inferring the distribution of three key parameters for quantifying the terrestrial biosphere.

\section{Proposed methodology} \label{sec:problem}

\subsection{Forward and inverse modeling}

\begin{figure}[h!]
\begin{center}
\begin{tikzpicture}[->,>=stealth',scale=0.9,transform shape,node distance=3cm,thick]
  \tikzstyle{every state}=[fill=gray!30,draw=none,text=black]
  \node[state] (RET) [fill=gray!60] {Inversion $g(e,\theta)$};
  \node[state] (RTM)  [above of=RET,fill=gray!60] {RTM $f(c,\phi)$};
  \node[state] (OBS)  [right of=RTM,fill=gray!80,yshift=-2cm] {Reflectances $e$};
  \node[state] (VARS) [left of=RTM,fill=gray!80,yshift=-2cm]  {Parameters $c$};
  \path[solid, bend left=25, thick, color=gray] (VARS) edge node {} (RTM);
  \path[solid, bend left=25, thick, color=gray] (RTM)  edge node[yshift=+0.5cm,xshift=+1cm] {forward problem} (OBS);
  \path[dashed, bend left=25, thick, color=gray] (OBS)   edge node {} (RET);
  \path[dashed, bend left=25, thick, color=gray] (RET)   edge node[yshift=-0.5cm,xshift=-1cm] {inverse problem} (VARS);
\end{tikzpicture}
\caption{The forward problem in Earth observation involves taking the system's structural state  
as an input, defining representative bio-geo-physical parameters (e.g. vegetation canopy or leaf characteristics), then propagating the solar energy through the atmosphere medium and producing a simulated at-sensor reflectance. The inverse problem involves performing inference over the forward model. We use $\cau$ to denote the model parameters or causes, $\f$ is an RTM or simulator, and $\e$ is the simulated output reflectance or effect. Both the forward RTM ${\bf f}$ and the inverse model ${\bf g}$ are  nonlinear functions parameterized by $\boldsymbol{\theta}$ and $\boldsymbol{\phi}$, respectively.
\label{forward_inverse}}
\end{center}
\end{figure}
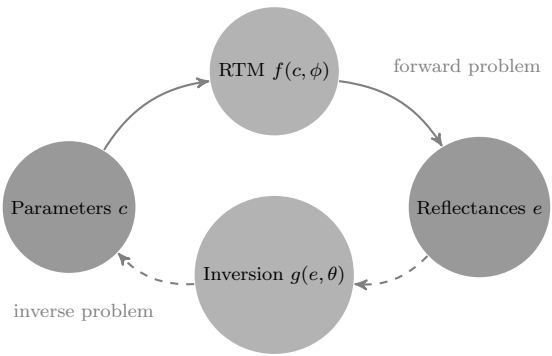

Notationally, an RTM $\f$ operating in {\em forward mode} generates a multidimensional {\em reflectance/radiance observation (or effect)} ${\bf e}\in\mathbb{R}^{D_e}$ as observed by the sensor given a multidimensional {\em parameter state vector (or cause)} ${\bf c}\in\mathbb{R}^{D_c}$, see Fig.~\ref{forward_inverse}. Running forward simulations yields a look-up-table (LUT) of input-output pairs, ${\mathcal D }=\{(\mathbf{c}_i,\mathbf{e}_i)\}_{i=1}^n$.  
Solving the {\em inverse problem} using machine learning implies learning the function $\g$ using ${\mathcal D }$, to return an estimate $\mathbf{c}_*$ each time a new satellite observation $\mathbf{e}_*$ is acquired. 

\subsection{Problem setting}

Notationally, let us consider then the vector of effects ${\bf e} \in\mathbb{R}^{D_e}$ and vector of causes ${\bf c} \in\mathcal{C}\subseteq\mathbb{R}^{D_c}$, an RTM model represents the underlying mapping from ${\bf c}$ to ${\bf e}$, that we denote as ${\bf f}({\bf c}): \mathbb{R}^{D_c} \rightarrow \mathbb{R}^{D_e}$.
The complete observation model is given by
\begin{eqnarray}\label{eq:generative}
    \e={\bf f}(\cau) + {\bm \epsilon}, ~~~~~  {\bm \epsilon} \sim \mathcal{N}({\bm \epsilon}|{\bf 0},\sigma^2\mathbf{I}),
\end{eqnarray}
where $\mathbf{I}$ is a unit $D_e \times D_e$ matrix. The observation model defines the likelihood function as
\begin{equation}\label{eq:llhood}
p({\bf e}|{\bf c})=\mathcal{N}({\e}|{\bf f}({\bf c}),\sigma^2\mathbf{I}).    
\end{equation}
Note that by fixing ${\bf c}$, the conditional probability $p({\bf e}|{\bf c})$ is Gaussian, but as a function of ${\bf c}$ the likelihood is a highly non-linear function due to the dependence on the RTM with the causes, i.e. ${\bf f}({\bf c})$. 
We assume an Gaussian prior over ${\bf c}$'s,  
\begin{eqnarray}
p(\cau) = \mathcal{N}(\cau|\mathbf{m}, \mathbf{S})\,,
\end{eqnarray}
where ${\bf m}\in \mathbb{R}^{D_c}$ and the $D_c\times D_c$ covariance matrix ${\bf S}$ are considered unknown.
The posterior density given the observed data $\e$ over the causes can be  expressed as 
\begin{eqnarray}
   p(\cau|\e)  &\propto& p(\e|\cau)p(\cau)= \mathcal{N}({\e}|{\bf f}({\bf c}),\sigma^2\mathbf{I})\mathcal{N}(\cau|\mathbf{m}, \mathbf{S})
\end{eqnarray}
Our goals are:  (a) learn the prior parameters, vector ${\bf m}$ and matrix ${\bf S}$, and (b)
 obtain an approximation of the posterior $p({\bf c}|{\bf e})$,  which serves as an inverse probabilistic mapping from ${\bf e}$ to ${\bf c}$. We assume that some set of data $\e$ is given. The two main ways of approaching this problem are a Variational inference (VI\footnote{Please note that we use \textit{VI} to abbreviate \textit{variational inference} as in most machine learning literature. This is not be confused with the abbreviation of \textit{vegetation indices} in the remote sensing literature.}) scheme on the one hand, and an expected maximization method on the other. For the VI method we follow the approach of Kingma and Welling \cite{kingma2013auto} and substitute the decoder network with the generative model of Eq.~\eqref{eq:generative}. For the MC-based approach we use MC Expectation Maximization \cite{wei1990monte}.

\subsection{Variational inference method}
The idea of variational inference is to optimize the parameters of a \textit{variational posterior} in order to come as close as possible to the \textit{true posterior}. Following \cite{kingma2013auto} we choose a Gaussian variational posterior,
\begin{eqnarray}\label{eq:varpostform}
    q(\cau|\e) = \mathcal{N}(\cau|{\bm \mu}_{\text{NN}}(\e), \mathbf{\Sigma}_{\text{NN}}(\e)),
\end{eqnarray}
where ${\bm \mu}_{\text{NN}}(\e)$ and $\mathbf{\Sigma}_{\text{NN}}(\e)$ are obtained by tuning a Neural Network (NN) with parameters ${\bm \phi}$. 
These parameters  ${\bm \phi}$ are also referred to as the variational parameters.
In order to tune the NN parameters  ${\bm \phi}$,  we minimize the Kullback-Leibler (KL) divergence between $q(\cau|\e)$ and the true posterior $p(\cau|\e)$, i.e.,
\begin{eqnarray}
    \mbox{KL}\left[q(\cau|\e)||p(\cau|\e)\right] &=& -\mathbb{E}_{q(\cau|\e)} \left[ \log \frac{p(\cau|\e)}{q(\cau|\e)} \right] \nonumber \\
    &=&-\mathbb{E}_{q(\cau|\e)} \left[ \log \frac{p(\cau,\e)}{q(\cau|\e)} - \log p(\e) \right] \nonumber \\
    &=& -\mathcal{L} + \log p(\e),
\end{eqnarray}
where we have used $\mathbb{E}_{q(\cau|\e)}[\log p(\e)]=\log p(\e)$.
Since $\log p(\e)$ is constant w.r.t. the variational parameters, in order to minimize the KL divergence, we have to maximize 
$$
\mathcal{L} = \mathbb{E}_{q(\cau|\e)} \left[ \log  \frac{p(\cau,\e)}{q(\cau|\e)} \right] = \mathbb{E}_{q(\cau|\e)} \left[ \log  \frac{p(\e|\cau)p(\cau)}{q(\cau|\e)} \right],
$$
 called the {\it Evidence Lower Bound} (ELBO). 
We can split the ELBO into two terms: the first one represents the expected log-likelihood with respect to the variational posterior and the second one is the KL divergence between the variational posterior and the prior, i.e.,
\begin{eqnarray}
    \mathcal{L} &=& \mathbb{E}_{q(\cau|\e)} \left[ \log  \frac{p(\e|\cau)p(\cau)}{q(\cau|\e)} \right], \nonumber \\
    &=& \mathbb{E}_{q(\cau|\e)} \left[ \log  p(\e|\cau)\right]  - \left[\frac{q(\e|\cau)}{p(\cau|\e)} \right], \nonumber \\
    &=& \mathbb{E}_{q(\cau|\e)} \left[ \log  p(\e|\cau)\right] - \mbox{KL}\left[q(\cau|\e)||p(\cau)\right].
\end{eqnarray}
As opposed to the approach in \cite{kingma2013auto} we place a deterministic forward model in stead of a decoder network and fix a low value of noise variance in the likelihood (Eq. \ref{eq:llhood}) in order to reflect the trust in the forward model. This approach is akin to that of \cite{mccarthy2017variational}.
In order to optimize this expression, we perform a Monte Carlo estimation of the expected value (i.e, the fist term) \cite{robert2013monte}.
The second term has a simple analytical form as it is the KL divergence between two Gaussians. 

Importantly, maximizing $\mathcal{L}$ with respect to $\bm{\phi}$ should make $\mbox{KL}\left[q(\cau|\e)||p(\cau|\e)\right]$ fairly small and hence  $\mathcal{L} \approx \log p(\mathbf{e})$. The maximization of $\mathcal{L}$ with respect to the prior parameters, $\bm{\theta}=\{\mathbf{m},\mathbf{S}\}$, is hence expected to maximize $\log p(\mathbf{e})$, which is the maximum likelihood principle for parameter estimation. In practice, we maximize $\mathcal{L}$ simultaneously with respect to $\bm{\theta}$ and $\bm{\phi}$. 

The previous approach can be easily extended to the case of having several observed data instances $\{\mathbf{e}_i\}_{i=1}^N$. In that case the objective is simply the sum of $\mathcal{L}_i$, for $i=1,\ldots,N$, where $\mathcal{L}_i$ is the lower bound corresponding to $\mathbf{e}_i$, \emph{i.e.}, the $i$-th data instance. This sum can be approximated using mini-batches and optimized using stochastic optimization techniques such as the ADAM algorithm \cite{kingma2014adam}. In this study, we use a mini-batch size of 1 for all experiments. For a proof of convergence of stochastic optimization see \cite{robbins1951stochastic}. The variational approach is expected to find reasonable values for the prior parameters $\bm{\theta}$, using approximate maximum likelihood estimation, and to provide a recognition model $q(\mathbf{c}|\mathbf{e})$ that can be used to infer the potential values of $\mathbf{c}$ given $\mathbf{e}$.

\subsection{Monte Carlo expectation maximization}

Another method which can be used to address the learning goals described in Section 
\ref{sec:problem}, i.e. to infer the prior parameters from the observed 
data, and to generate samples from the posterior distribution $p(\mathbf{c}|\mathbf{e})$, is 
the Monte Carlo Expectation Maximization (MCEM) method \cite{wei1990monte}.

We begin by briefly describing the  Expectation Maximization (EM) algorithm,
which can be used to maximize the likelihood 
function in models that involve latent variables \cite{dempster1977maximum}. This is precisely
the scenario considered in Section \ref{sec:problem}. Namely, given some observed data 
$\{\mathbf{e}_i\}_{i=1}^N$, we would like to maximize
\begin{align}
	\prod_{i=1}^N p(\mathbf{e}_i|{\bm \theta}) = \prod_{i=1}^N \int_{\mathcal{C}} p(\mathbf{e}_i|\mathbf{c}_i) 
	p(\mathbf{c}_i|{\bm \theta}) d \mathbf{c}_i\,, \label{eq:likelihood_em}
\end{align}
as a function of the prior parameters ${\bm \theta}=\{\mathbf{m},\mathbf{S}\}$. 
Direct optimization of (\ref{eq:likelihood_em}) is intractable, since we 
cannot marginalize the latent variables $\mathbf{c}_i$. The EM algorithm uses 
the fact that the complete likelihood function $p(\mathbf{e}_i,\mathbf{c}_i|{\bm \theta}) = 
p(\mathbf{e}_i|\mathbf{c}_i) p(\mathbf{c}_i|{\bm \theta})$ is tractable. Consider the following decomposition of the logarithm of (\ref{eq:likelihood_em})
\begin{align}
\log \sum_{i=1}^N p(\mathbf{e}_i|{\bm \theta}) &= \sum_{i=1}^N \mathcal{L}(q_i,{\bm \theta}) + \text{KL}(q_i||p_i)\,,
	\label{eq:log_likelihood_em}
\end{align}
where we have introduced an approximate distribution $q_i(\mathbf{c}_i)$ and
\begin{align}
	\mathcal{L}(q_i,{\bm \theta}) &= \int_{\mathcal{C}} q_i(\mathbf{c}_i) \log \frac{p(\mathbf{e}_i,\mathbf{c}_i|{\bm \theta})}
	{q_i(\mathbf{c}_i)} d \mathbf{c}_i\,, \\
	\text{KL}(q_i||p_i) &= - \int_{\mathcal{C}} q_i(\mathbf{c}_i) \log 
	\frac{p(\mathbf{c}_i|\mathbf{e}_i,{\bm \theta})}{q_i(\mathbf{c}_i)} d \mathbf{c}_i\,. \label{eq:kl_em}
\end{align}
Note that (\ref{eq:kl_em}) is the Kullback Leibler divergence between $q_i$ and the exact posterior $p(\mathbf{c}_i|\mathbf{e}_i,{\bm \theta})$
for the instance $\mathbf{e}_i$.

The EM algorithm maximizes (\ref{eq:log_likelihood_em}) in a two stage iterative process. Assume the current
parameter vector is ${\bm \theta}^\text{old}$. In the E step, the lower bound $\sum_{i=1}^N \mathcal{L}(q_i,{\bm \theta}^\text{old})$
is maximized with respect to each $q_i$ assuming ${\bm \theta}^\text{old}$ to be fixed. 
Because $\sum_{i=1}^N \log p(\mathbf{e}_i|{\bm \theta})$ does not depend on each $q_i$, the solution to this problem 
consists in setting each $q_i(\mathbf{c}_i)$ equal to $p(\mathbf{c}_i|\mathbf{e}_i,{\bm \theta}^\text{old})$, minimizing
$\text{KL}(q_i||p_i)$ in consequence. In the subsequent M step, each $q_i(\mathbf{c}_i)$ is held fixed, 
and $\sum_{i=1}^N \mathcal{L}(q_i,{\bm \theta}^\text{old})$ is maximized with respect to ${\bm \theta}$, to give new prior parameters ${\bm \theta}^\text{new}$. This will cause the lower bound $\sum_{i=1}^N \mathcal{L}(q_i,{\bm \theta}^\text{old})$ 
to increase, which will in turn increase the log-likelihood $\sum_{i=1}^N \log p(\mathbf{e}_i|{\bm \theta})$. 
Critically, $q_i(\mathbf{c}_i)$ will be computed in this step using ${\bm \theta}^\text{old}$, which is fixed. Therefore,
the only required integral to evaluate in the M step is
\begin{align}
	\mathcal{L}(q_i,{\bm \theta}) &= \int_{\mathcal{C}} q_i(\mathbf{c}_i) \log p(\mathbf{c}_i|{\bm \theta}) d \mathbf{c}_i + \text{const.}
	\label{eq:intractable}
\end{align}
A difficulty, however, is that the posterior $p(\mathbf{c}_i|\mathbf{e}_i,{\bm \theta}^\text{old})$ is intractable, which makes computing $q_i$ and hence the integral in (\ref{eq:intractable}) challenging. Monte Carlo EM (MCEM), provides a
solution to this problem \cite{wei1990monte}. The intractable integral in (\ref{eq:intractable}) is 
simply approximated by a Monte Carlo average over several samples drawn from $q_i$. Namely, 
\begin{align}
\mathcal{L}(q_i,{\bm \theta}) & \approx \frac{1}{S} \sum_{s=1}^S  \log p(\mathbf{c}_i^s|{\bm \theta}) + \text{const.}
	\,,
	\label{eq:monte_carlo_approx}
\end{align}
where $\mathbf{c}_i^s$ has been generated from $q_i$ and $S$ is the number of generated samples. 
The convergence properties of MCEM are analyzed in \cite{neath2013convergence}.

Recall that the approximate distribution $q_i$ is targeting the exact posterior 
$p(\mathbf{c}_i|\mathbf{e}_i,{\bm \theta}^\text{old})$. So ideally, we should generate the 
samples $\mathbf{c}_i^s$ from the exact posterior. For this, we use Hamilton Monte 
Carlo (HMC) \cite{neal2011mcmc} as in \cite{kingma2013auto}. HMC is a Markov chain Monte Carlo (MCMC) method that can be used to generate (correlated) samples from some target 
distribution \cite{MartinoMh17}. More specifically, a Markov chain is generated whose stationary distribution coincides
with the target distribution. By running the Markov chain for a sufficiently large number
of steps one can obtain an approximate independent sample from $p(\mathbf{c}_i|\mathbf{e}_i,{\bm \theta}^\text{old})$. 
HMC has the advantage, when well-tuned,  reduces  substantially the correlation among samples \cite{MartinoMh17}. For this, it simulates a dynamical system that 
uses information about the gradient of the posterior, \emph{i.e.}, $\nabla_{\mathbf{c}_i} \log 
p(\mathbf{c}_i|\mathbf{e}_i,{\bm \theta}^\text{old})$, to sample from regions of high posterior probability.
In our implementation of MCEM, the HMC procedure consists of 20 leapfrog steps with small step-size (\emph{i.e.}, $5 \times 10^{-4}$) which guarantees that the acceptance rate is high enough. In 
practice, we only use just one sample to approximate (\ref{eq:monte_carlo_approx}). Each time, the 
Markov chain is initialized at the mode of the posterior distribution, which is found using 
quasi-newton optimization methods (\emph{i.e.}, L-BFGS).
Of course, after optimizing the prior parameters ${\bm \theta}$ using MCEM, HMC can be used to generate
samples from the approximate posterior distribution $\widehat{p}(\mathbf{c}|\mathbf{e},{\bm \theta}) \propto p(\e|\cau)\widehat{p}_{{\bm \theta}}(\cau)$.


\subsection{Important considerations}

Note that both, the variational and MCEM methods, provide an estimation of the parameters ${\bm \theta}$ of the prior. Thus, we obtain a Gaussian approximation of the prior, which is denoted here as  $\widehat{p}_{{\bm \theta}}(\cau)$. Therefore, both techniques provide the following posterior approximation 
\begin{align}
    \widehat{p}(\cau|\e,{\bm \theta})\propto p(\e|\cau)\widehat{p}_{{\bm \theta}}(\cau).
\end{align}
However, the variational algorithm provides another posterior approximation given  in Eq.\eqref{eq:varpostform}, i.e., 
\begin{eqnarray}\label{eq:varpostform2}
    q(\cau|\e) = \mathcal{N}(\cau|{\bm \mu}_{\text{NN}}(\e), \mathbf{\Sigma}_{\text{NN}}(\e)),
\end{eqnarray}
which yields an important advantage  with respect the previous one: given one $\e$, using $q(\cau|\e)$ we can easily, and at low computational cost, produce a predictive mean ${\bm \mu}_{\text{NN}}(\e)$ and covariance $\mathbf{\Sigma}_{\text{NN}}(\e)$. The approximation $ \widehat{p}(\cau|\e,{\bm \theta})$ on the other hand would require the use of additional Monte Carlo schemes for obtaining a predictive mean and variance, for each new observation vector \e.
Another advantage of the variational approach is the computational speed compared to the MCEM method. However, one advantage of the MCEM scheme is that it can directly handle more practical scenarios (\emph{e.g.}, problems involving multiple posterior modes, heavy tailed distributions,  etc.) leading to better performance in terms of smaller error in the parameter estimation of the prior.
The variational approach described here would require a different and more general derivation for addressing these scenarios, see \emph{e.g.}  \cite{mescheder2017adversarial}. These features of each method are confirmed by the results obtained in our experiments. Implementations of the two approaches can be found at \href{https://github.com/dhsvendsen/RTM_VI_MCEM_INFERENCE}{github.com/dhsvendsen/rtm\_vi\_mcem\_inference}.

\section{Experiments} \label{see:intro}
We illustrate the strengths and weaknesses of the two approaches, first by means of informative toy experiments: One that studies the computational efficiency of the respective methods, and another which analyzes their ability to handle forward models leading to multimodal posteriors. Following this, we show how these approaches can be used to perform inference over biophysical parameters using an RTM as the forward model.

\subsection{On the computational efficiency}
In order to analyze the computational efficiency of the two approaches, we consider a simple forward model
$$
{\bf f}(\cau)=f(c_1,c_2) = [2c_1, 2c_2],
$$
for which both approaches converge to the true values of the parameters of the prior. We draw the training data $\{{\bf c}_n\}_{n=1}^N$ from the prior
$$p(\cau) = \mathcal{N}\left( \left[ 
\begin{matrix} 4 \\ 6 \end{matrix} \right] \,, ~
\left[ \begin{matrix} 1 & 0.6 \\ 0.6 & 1 \end{matrix} \right]
\right),
$$
and pass it through the nonlinear mapping ${\bf f}$ in order to generate the training data $\{\e_n\}_{n=1}^N$. Datasets of several sizes $N = \{50, 500, 1000, 2000\}$ are used for training the models. The model likelihood noise is in all experiments fixed at a negligible value, with $\sigma^2 = 10^{-7}$, in order to reflect the trust in the knowledge encoded in the RTMs. 

\begin{figure}[h!]
    \centering
    \includegraphics[width=1\textwidth]{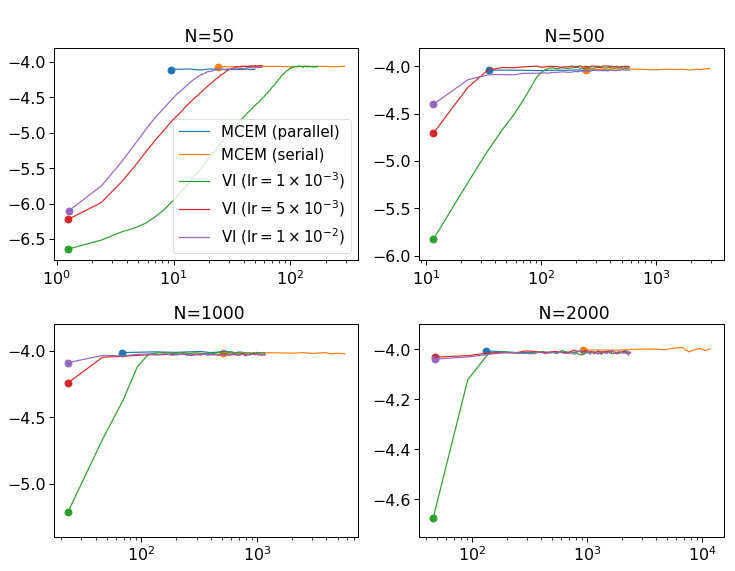}
    \caption{Marginal log-likelihood of test dataset as a function of training time for different inference methods. Four different sizes of training datasets are used, showing that the VI method is computationally more efficient than the MCEM method for larger datasets.}
    \label{fig:computeff}
\end{figure}

In Fig. \ref{fig:computeff} we plot an estimate of the average log marginal likelihood of each method on a test dataset as a function of training time (averaged over 40 repetitions). The marginal likelihood is computed using the estimator described in Appendix \ref{sec:estimator}. Observing the test log-likelihood, which is computed after each epoch, we see that the MCEM method convergences after 1 epoch (one iteration of the E and M steps). With a training dataset of 50 points, each epoch of training is sufficiently fast that a parallelized version of the MCEM method (in which each E step is done in parallel) converges faster than the VI method. For the non-parallelized algorithm, this is not the case. For larger datasets, VI converges before the completion of 1 epoch of the MCEM algorithm. Since this is a simple toy problem, larger learning rates can be used in the VI method, leading to earlier convergence (just after 1 epoch) for datasets of 1000 and 2000 points. We can conclude from these experiments that the VI approach (as a consequence of stochastic optimization) has a better scaling properties with respect to the dataset size than MCEM.

\subsection{Dealing with multimodal posteriors}

We have seen that when faced with sufficiently large amounts of training data, variational inference performs faster than Monte Carlo sampling methods. However, since the form of the variational posterior assumed in Eq.~\eqref{eq:varpostform} is unimodal, we cannot expect it to be able to capture any multimodality in the true posterior. Consider for instance the forward mapping (with $\cau=[c_1,c_2]$),
$${\bf f}(\cau) = [c_1^2, c_1 c_2].$$
For a given observed $\e=[e_1,e_2] \geq [0,0]$ there will always be two possible solutions, namely $\cau^{(1)}=[\sqrt{e_1}, e_2/\sqrt{e_1}~]$ and $\cau^{(2)}=[-\sqrt{e_1}, -e_2/\sqrt{e_1}~]$ making the posterior inherently multimodal.
As stated in the previous sections, we consider ${\bm \epsilon} \sim \mathcal{N}({\bm \epsilon}|{\bf 0},10^{-7}\mathbf{I})$. In this example, the prior density is Gaussian with parameters
$$
p(\cau) = \mathcal{N}\left( \left[ 
\begin{matrix} 1 \\ 2 \end{matrix} \right] \,, 
\left[ \begin{matrix} 1 & 0.6 \\ 0.6 & 1 \end{matrix} \right]
\right),
$$
from which $500$ samples are drawn and passed through $\f$ to generate the training dataset.
\begin{figure}[h!]
    \centering
    \includegraphics[width=0.9\textwidth]{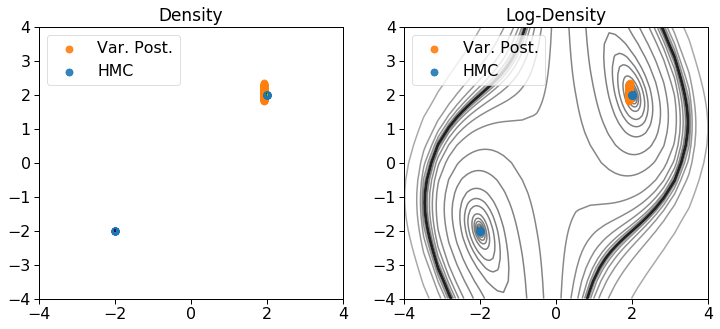}
    \caption{Contour plots of samples from the true posterior conditioned on the observation $\e=[4,\, 4]^\top$. HMC samples using the prior parameters learned by the MCEM method shown in blue, and samples from the learned variational posterior in  orange. The density (left) is so sharply peaked around the two modes that it is more informative to study the log-density (right).}
    \label{fig:multimode_posterior}
\end{figure}

 In the process of maximizing the ELBO, the expected log-likelihood with respect to the variational posterior is computed. We can see from Fig. \ref{fig:multimode_posterior}, however, that the variational posterior, upon convergence, only captures the positive mode at $\cau'=[2,\, 2]^\top$ of the true posterior given the observation $\e'=[4,\, 4]^\top$. On the other hand, the MCEM algorithm computes the expected complete log-likelihood with respect to the true posterior as approximated with HMC. As opposed to the variational posterior, HMC does manage to capture both the modes of the true posterior as shown in Fig. \ref{fig:multimode_posterior}. The learning algorithm of the MCEM method is therefore more likely to converge to the true parameters of the prior if the posterior is multimodal.

We can see the inability of the variational method to capture the multimodality of the problem from the results of the converged methods given in Table \ref{tab:toy2}. The fitted parameters of the prior are far from the true ones when compared to the results of the MCEM method which as also reflected in the KL divergence between the fitted and true prior distributions. Multimodality such as this is likely to be observed in the remote sensing experiment latter, as it has been remarked before that different configurations of inputs can lead to the same output making it an \textit{ill-posed inversion problem} \cite{gomez2016efficient}.

\begin{table}[h!]
  \centering
\begin{tabular}{|>{\centering\arraybackslash}p{.12\textwidth}|*2{>{\centering\arraybackslash}p{.25\textwidth}|}}
\hline
Method & \textbf{VI} & \textbf{MCEM} \\
\hline
\[ \mbox{Mean} \] &\vspace{-0.5cm} {  \[\left[  \begin{array}{c} 1.20 \\ 1.76 \end{array} \right]\] }  &\vspace{-0.5cm}
{  \[\left[  \begin{array}{c} 1.027 \\ 2.061 \end{array} \right]\] } \\
\hline
\[ \mbox{Covariance} \]  &\vspace{-0.5cm} {  \[\left[  \begin{array}{cc} 0.619 & 0.682 \\ 0.682 & 1.600 \end{array}\right]\] } & 
\vspace{-0.5cm} {  \[\left[  \begin{array}{cc} 1.001 & 0.617 \\ 0.617 & 0.945 \end{array}\right]\] } \\
\hline
$D_{KL}$ & 0.315 & 0.00581 \\
\hline
\end{tabular}
    \caption{Comparison of methods for inference on a forward model which leads to a bimodal posterior. The first and second columns show the estimated mean vector and covariance matrix of the prior. The third column shows the KL divergence between the fitted and the true prior.}
    \label{tab:toy2}
\end{table}

\subsection{PROSAIL experiment}

We now turn to inference in a remote sensing setting using one of the most widely used RTM over the last {almost three} decades in the field as our physical forward model \cite{jacquemoud2009prospect+}. PROSAIL is a canopy reflectance model which allows us to relate fundamental vegetation canopy {properties}, such as, the Leaf Area Index (LAI), and leaf chemical and structural properties, to the scene reflectance for {a given set of} illumination and sensor (observation) geometry conditions \cite{liang2005quantitative}. To perform its simulations, PROSAIL combines two sub-models: PROSPECT \cite{feret2008prospect}, which models the optical properties of the leaves; and SAIL \cite{verhoef1984light}, which models bidirectional reflectances considering the scattering by arbitrarily inclined canopy leaves in a turbid medium \cite{fang2019overview}. This combination of models requires the following set of input parameters:
\begin{itemize}
    \item[1)]  A set of leaf optical properties (PROSPECT), given by the mesophyll structural parameter (N), leaf chlorophyll (Chl), dry matter (Cm), water (Cw), carotenoid (Car) and brown pigment (Cbr) contents.
    \item[2)] A set of canopy level and geometry characteristics (SAIL), determined by leaf area index (LAI), the average leaf angle inclination (ALA), the hot-spot parameter (Hotspot), the solar zenith angle ($\theta_s$), view zenith angle ($\theta_v$), and the relative azimuth angle between both angles ($\Delta\Theta$).
\end{itemize} 
We consider PROSAIL for simulating Landsat-8 spectra. This satellite has been widely used in many applications such as cryosphere monitoring, aquatic science and surface water mapping, and vegetation monitoring \cite{wulder2019remote}. Landsat 8's Operational Land Imager (OLI) includes nine spectral bands with wavelengths ranging from $0.433 \mu m$ to $1.390 \mu m$, leaving us with an output-dimension of $D_e=9$ for our problem. 
In our experimental setup, we have chosen to work with the most relevant leaf-level parameters to monitor vegetation status and functioning included in PROSAIL, namely Cw, Cm and Chl, resulting in an input dimension of $D_c=3$. The remaining parameters were set constant during our experiments and their values were obtained from previous studies \cite{amogape20} to be representative of realistic cases. Their values can be found in Table \ref{tab:provals}.

\begin{table}[h!]
\centering
\caption{Characteristics of the simulations using the PROSAIL model. \label{tab:provals}}
\vspace{0.0cm} 
\begin{tabular}{|c|c|c|c|c|c|c|}
\cline{1-4} 
\multirow{2}{*}{\emph{Leaf}}   & N & \multicolumn{1}{c|}{Car} & \multicolumn{1}{c|}{Cbr} &  \multicolumn{1}{c}{ }  & \multicolumn{1}{c}{ } & \multicolumn{1}{c}{ } \\ \cline{2-4} 
                              & 1.5 & \multicolumn{1}{c|}{  8 g/cm$^2$ } & \multicolumn{1}{c|}{ 0 } & \multicolumn{1}{c}{ } &  \multicolumn{1}{c}{ } & \multicolumn{1}{c}{ } \\ \Xhline{2\arrayrulewidth}
\multirow{2}{*}{\emph{Canopy}} & ALA & Hotspot & $\theta_s$ & $\theta_\nu$ & $\Delta\Theta$ & LAI \\ \cline{2-7} 
                              &  Spherical & 0.01 & 30$^\circ$ & 10$^\circ$ & 0 & 4  \\ \hline
\end{tabular}
\end{table}

Constraining the radiative transfer models with realistic and representative distributions of their inputs is a key part of the RTM inversion process. To facilitate this, in this work we relied on the largest global plant traits database available, the \href{https://www.try-db.org/}{TRY} database \cite{kattge2011try,kattge2020try}, which contains thousands of leaf data records measured at unprecedented spatial and climatological coverage. Using these data we computed the following empirical mean vector and covariance matrix which was used to sample 2000 values of $\cau$ and pass them through PROSAIL to generate the training data. The empirical mean and covariance (to be compared with the results in Table \ref{tab:vi_vs_mcem}) of the samples are
$$
\hat{\mathbf{m}} =  \left[ 
\begin{matrix} 9.76e^{\smin3} \\ 1.77e^{\smin2} \\ 46.2 \end{matrix} \right] \, ,~ \hat{\mathbf{S}} = 
\left[ \begin{matrix} 6.42e^{\smin5} & 5.06e^{\smin5} & 3.68e^{\smin2} \\ 5.06e^{\smin5} & 1.34e^{\smin4} &\smin2.86e^{\smin3} \\ 3.68e^{\smin2} &\smin2.86e^{\smin3} & 288 \end{matrix} \right].
$$
The units of the parameters are g/cm$^2$ for Cm and Cw, and $\mu$g/cm$^2$ for Chl respectively. Note that the ground truth prior estimated from the TRY database has some probability density in the negative region of parameter space. This is not physically meaningful, but serves the point of illustrating the capabilities of the inference methods. We alter PROSAIL so that it sets every negative parameter to 0 before mapping into spectral space to get a modified likelihood that will lead to more multimodality (since all negative values in $\cau$ will be mapped into the same value, i.e. 0, and then through PROSAIL into a spectrum).

\begin{figure}[h!]
    \centering
    \includegraphics[width=\textwidth]{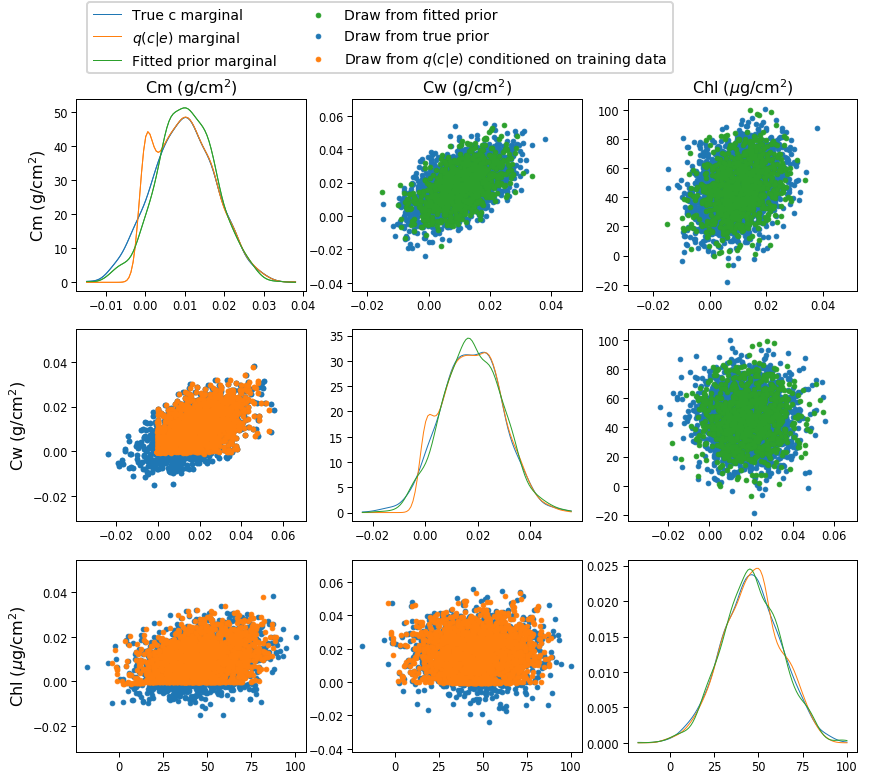}
    \caption{Results of the variational approach to inference over PROSAIL. The blue points are $\cau$'s from the training set, while the green points are draws from the fitted prior. The orange points are draws from the variational posterior conditioned on the training $\e$'s. The diagonal shows KDE plots of $\cau$ using samples from the ground truth prior (blue), the variatonal posterior conditioned on training data (orange) and the fitted prior (green).}
    \label{fig:3D_overview}
\end{figure}

The results of the variational approach to inference over PROSAIL are summarized in Fig. \ref{fig:3D_overview}. We see that the parameters of the prior are fitted well, which can also be confirmed in Table \ref{tab:vi_vs_mcem} quantitatively, even though the variational posterior is not able to produce predictive means in the negative domain. It is interesting to note that the modification of PROSAIL to truncate negative data, which leads to multimodality, does not prevent the variational approach from estimating the parameters of the prior well.

Nevertheless, the MCEM method is somewhat more accurate than the VI method, obtaining a KL divergence with to the true prior of $1.23\times10^{-2}$ compared to $2.08\times10^{-2}$ obtained using the VI approach. This is to be expected since, as we have seen, the MCEM approach handles multimodality better. We especially foresee a clear difference in results in future work the LAI variable which is difficult to estimate due to its multimodal posterior distribution as pointed out elsewhere \cite{gomez2016efficient}.

\begin{table}[h!]
    \centering

\begin{tabular}{|>{\centering\arraybackslash}p{.12\textwidth}|*2{>{\centering\arraybackslash}p{.32\textwidth}|}}
\hline
Method & \textbf{VI} & \textbf{MCEM} \\
\hline
\[ \mbox{Mean} \] &\vspace{-0.5cm} {  \[\left[ \begin{array}{c} 1.02e^{\smin2} \\ 1.80e^{\smin2} \\ 45.9  \end{array} \right]\] } &\vspace{-0.5cm}
{  \[\left[  \begin{array}{c} 1.01e^{\smin2} \\ 1.81e^{\smin2} \\ 46.6  \end{array} \right]\] } \\
\hline
\[ \mbox{Covariance} \] &\vspace{-0.5cm} {  \[\left[  \begin{array}{ccc} 5.32e^{\smin5} & 4.31e^{\smin5} & 3.52e^{\smin2} \\ 4.31e^{\smin5} & 1.19e^{\smin4} & 2.18e^{\smin3} \\ 3.52e^{\smin2} & 2.18e^{\smin3} & 280 \end{array}\right]\] } & 
\vspace{-0.5cm} {  \[\left[  \begin{array}{ccc} 5.49e^{\smin5} & 4.45e^{\smin5} & 3.31e^{\smin2} \\ 4.45e^{\smin5} & 1.25e^{\smin4} &\smin1.74e^{\smin3} \\ 3.31e^{\smin2} &\smin1.74e^{\smin3} & 292 \end{array}\right]\] } \\
\hline
$D_{KL}$ & 0.0208 & 0.0123 \\
\hline
\end{tabular}

    \caption{Comparison of methods for inference on biophysical parameters using a radiative transfer forward model. The first and second columns show the mean vector and covariance matrix respectively of the true and the estimated prior over the causes, using $e=\times 10$ notation for space. The third column shows the Kullback-Leibler divergence between the fitted and the true prior.}
    \label{tab:vi_vs_mcem}
\end{table}{}

\begin{figure}[h!]
    \centering
    \includegraphics[width=0.95\textwidth]{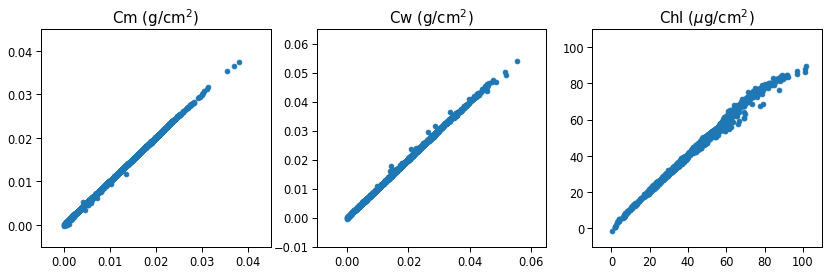}
    \caption{True values of RTM parameters in test dataset versus mean of variational posterior conditioned on spectra in test dataset. The trained encoder network can thus be used as an effective predictive model}
    \label{fig:3D_pred}
\end{figure}

Once the VI method has converged, the neural network which parameterizes the variational posterior can be used as a fast inverse model that maps from observed satellite spectra to biophysical variables. Using the mean outputs that model the mean value of the variational posterior we can obtain good predictive accuracy on a test set as shown in the scatter plots of Fig. \ref{fig:3D_pred}. Despite the promising results, it is very important to note that we run our experiments using a simplified PROSAIL configuration, keeping some of the input parameters static (see Table \ref{tab:provals}) and that results can vary greatly in more realistic modeling scenarios.

\section{Discussion and conclusions} \label{see:intro}

In this work, we approached the long-standing inverse problem in remote sensing of estimating biophysical parameters from observational reflectances.
Unlike previous works, we focus on estimating not only the particular parameter point estimates but its full multivariate distribution. We evaluated two different approximations that include an RTM forward model to enforce the inverse estimations to be physically consistent.

Both proposed techniques have different advantages and shortcomings that we illustrated with toy examples and with simulations from the PROSAIL RTM. 
The MCEM-based approach admits more flexible models while the VAE is computationally more efficient. For instance, while MCEM deals easily with multimodal distributions, this is a challenge for VAE. On the other hand, the convergence time of VAE is orders of magnitude faster depending on the problem.  Moreover, the VAE scheme provides a posterior approximation, with a predictive mean and a covariance matrix, implicitly defined by the trained neural network that can be readily evaluated.
The experiment involving PROSAIL shows that, while the accuracy of the VAE and MCEM are deemed similar, the computational simplicity of the VAE approach is critical in this problem. Note that including the RTM PROSAIL in the forward-inverse modeling loop increases the time of computation and combining it with MCEM makes it unfeasible especially for large data sets.

We anticipate a wide interest in these techniques for inferring the parameter densities from simulations and then, as further work, from observational satellite data. This will require more accurate and realistic priors; for this we plan to explore mixtures of Gaussians for modeling the prior of the causes as a generalization of the simplified Gaussian model assumed in this work. Likewise,  more sophisticated computational methods and variational approaches (\emph{e.g.}, \cite{Bugallo2015,MartinoMh17,mescheder2017adversarial}) could be explored in the future. 

Finally, we are well aware of the fact that this problem is ubiquitous in other domains of Earth observation and geosciences, and may have implications in climate science too. Inferring parameters is a transversal important topic, not only attached to terrestrial biosphere processes but to the atmosphere, cryosphere and the ocean modeling too.  For instance parametrization of small-scale processes such as clouds or biological processes (that are important at the land surface for the exchange of energy and carbon) cannot be explicitly resolved. In this context, learning appropriate parametrizations directly from data may reduce the sources of uncertainties in current models, eventually leading to a deadlock in climate modeling.


\appendix
\section{Marginal likelihood Estimation by Reverse Importance Sampling } \label{sec:estimator}

In order to evaluate the performance of the methods described in the paper, we use a estimator of marginal likelihood on a test-dataset. More precisely, we use 
the Reverse Importance Sampling (RIS) estimator  \cite{Llorente19ml} described below:

\begin{itemize}
    \item[1.] Sample $L$ values $\{\cau_l\}_{l=1}^L$ from the posterior with an MCMC-method. We use Hamiltonian Monte Carlo.
    \item[2.] Fit at density estimator $q(\cau)$ to the samples $\{\cau_l\}_{l=1}^L$. In this work we fit a Gaussian mixture model, doing cross validation in order to find the best number of components.
    \item[3.] Sample $M$ new values $\{\cau_m\}_{l=1}^M$ from the posterior to be inserted in the following estimator:
\end{itemize}
$$ p(\e) \simeq \left( \frac{1}{M}\sum_{m=1}^M \frac{q(\cau_m)}{p(\cau_m) p(\e|\cau_m)} \right)^{-1} ~~ \mbox{where} ~~ \cau_m \sim p(\cau | \e).$$
For the proof and more details see \cite{Llorente19ml}. It is important to remark that the function $q({\bf c})$ must be a valid probability density for which there are several possible choices. If one chooses $q({\bf c})=p({\bf c})$, RIS becomes the so-called harmonic mean estimator (this name is due to the fact that the corresponding estimator is the harmonic mean of the likelihood values). However, it has been shown that this does not lead to a good estimator. It is possible to show  that, in order to ensure finite variance of the resulting estimator, the density $q({\bf c})$ should have equal or lighter tails  than  the posterior $p({\bf c}|\e)$ (e.g., see first numerical example in \cite{Llorente19ml}). Gaussian mixture approximations and kernel density estimators of $p({\bf c}|\e)$ are suitable choices for $q({\bf c})$. Different  alternative estimators of the marginal likelihood are possible mixing MCMC and importance sampling schemes (see \cite{Llorente19ml,LAIS}).

\end{document}